%% file: main.tex
\documentclass[sigconf]{acmart}

\copyrightyear{2022}
\acmYear{2022}
\setcopyright{acmlicensed}
\acmConference[SIGIR '22] {Proceedings of the 45th International ACM SIGIR Conference on Research and Development in Information Retrieval}{July 11--15, 2022}{Madrid, Spain.}
\acmBooktitle{Proceedings of the 45th International ACM SIGIR Conference on Research and Development in Information Retrieval (SIGIR '22), July 11--15, 2022, Madrid, Spain}
\acmPrice{15.00}
\acmISBN{978-1-4503-8732-3/22/07}
\acmDOI{10.1145/3477495.3531972}
\usepackage{balance}
\usepackage{enumerate}
\usepackage{amsthm}
\usepackage{MnSymbol}
\usepackage{subfigure}
\usepackage{color}
\usepackage[normalem]{ulem}
\usepackage[export]{adjustbox}
\newcommand{\eqaref}[1]{Equation\ \ref{#1}}
\newcommand{\figref}[1]{Figure\ \ref{#1}}
\newcommand{\tabref}[1]{Table\ \ref{#1}}
\newcommand{\thmref}[1]{Theorem\ \ref{#1}}
\newtheorem{thm}{Theorem}

\usepackage[normalem]{ulem}
\useunder{\uline}{\ul}{}
\AtBeginDocument{%
  \providecommand\BibTeX{{%
    \normalfont B\kern-0.5em{\scshape i\kern-0.25em b}\kern-0.8em\TeX}}}

\begin{document}
\fancyhead{}

\title{ESCM$^2$: Entire Space Counterfactual Multi-Task Model for Post-Click Conversion Rate Estimation}

\author{Hao Wang}
\affiliation{
  \institution{Ant Group}
  \city{Hangzhou}
  \country{China}}
\email{zhiqiang.wh@antgroup.com	}

\author{Tai-Wei Chang}
\affiliation{
  \institution{Ant Group}
  \city{Hangzhou}
  \country{China}}
\email{taiwei.twc@antgroup.com}

\author{Tianqiao Liu}
\affiliation{
  \institution{Ant Group}
  \city{Hangzhou}
  \country{China}}
\email{liutianqiao.ltq@antgroup.com}

\author{Jianmin Huang}
\affiliation{
  \institution{Ant Group}
  \city{Hangzhou}
  \country{China}}
\email{caiqiong@antgroup.com}

\author{Zhichao Chen}
\affiliation{
  \institution{Ant Group}
  \city{Hangzhou}
  \country{China}}
\email{czc347886@antgroup.com}

\author{Chao Yu}
\affiliation{
  \institution{Ant Group}
  \city{Hangzhou}
  \country{China}}
\email{tianjing.yc@antgroup.com	}

\author{Ruopeng Li}
\affiliation{
  \institution{Ant Group}
  \city{Hangzhou}
  \country{China}}
  \email{ruopeng.lrp@antfin.com}
\author{Wei Chu}
\affiliation{
  \institution{Ant Group}
  \city{Hangzhou}
  \country{China}}
\email{weichu.cw@antgroup.com} 
\renewcommand{\shortauthors}{Hao Wang, Tai-Wei Chang, Tianqiao Liu, et al.}
\newcommand{\ie}{\textit{i}.\textit{e}. }
\newcommand{\eg}{\textit{e}.\textit{g}. }

\begin{abstract}
Accurate estimation of post-click conversion rate is critical for building recommender systems, which has long been confronted with sample selection bias and data sparsity issues. Methods in the Entire Space Multi-task Model (ESMM) family leverage the sequential pattern of user actions, \ie $impression\rightarrow click \rightarrow conversion$ to address data sparsity issue. However, they still fail to ensure the unbiasedness of CVR estimates. In this paper, we theoretically demonstrate that ESMM suffers from the following two problems: (1) Inherent Estimation Bias (IEB) for CVR estimation, where the CVR estimate is inherently higher than the ground truth; (2) Potential Independence Priority (PIP) for CTCVR estimation, where ESMM might overlook the causality from click to conversion. To this end, we devise a principled approach named Entire Space Counterfactual Multi-task Modelling (ESCM$^2$), which employs a counterfactual risk miminizer as a regularizer in ESMM to address both IEB and PIP issues simultaneously. Extensive experiments on offline datasets and online environments demonstrate that our proposed ESCM$^2$ can largely mitigate the inherent IEB and PIP issues and achieve better performance than baseline models.
\end{abstract}

\begin{CCSXML}
<ccs2012>
   <concept>
       <concept_id>10002951.10003317.10003347.10003350</concept_id>
       <concept_desc>Information systems~Recommender systems</concept_desc>
       <concept_significance>500</concept_significance>
       </concept>
   <concept>
       <concept_id>10010147.10010257.10010258.10010262</concept_id>
       <concept_desc>Computing methodologies~Multi-task learning</concept_desc>
       <concept_significance>300</concept_significance>
       </concept>
   <concept>
       <concept_id>10010147.10010178.10010187.10010192</concept_id>
       <concept_desc>Computing methodologies~Causal reasoning and diagnostics</concept_desc>
       <concept_significance>300</concept_significance>
       </concept>
 </ccs2012>
\end{CCSXML}

\ccsdesc[500]{Information systems~Recommender systems}
\ccsdesc[300]{Computing methodologies~Multi-task learning}
\ccsdesc[300]{Computing methodologies~Causal reasoning and diagnostics}

\keywords{Recommender System; Entire Space Multi-task Learning; Selection Bias; Post-click Conversion Rate Estimation}

\maketitle
\input{1_Introduction}
\input{2_preliminary}
\input{3_discussion}
\input{4_proposed}

\input{5_experiment}
\input{6_related}

\section{Conclusion and future work}
Due to the effectiveness of modeling associations between tasks, ESMM dominates many large-scale business scenarios. However, it still suffers from inherent estimation bias for its CVR estimation and potential independence priority for its CTCVR estimation. A principled approach named ESCM$^2$ is devised to augment ESMM with counterfactual regularization. Extensive experiments demonstrate that ESCM$^2$ can largely mitigate IEB and PIP, and achieve better performance compared with other baseline models.

There are two directions of future work that we intend to pursue. The first direction is to extend our causal framework to model more complex sequential user behaviors. Existing approaches \cite{esm2,gmcm}, in accordance with ESMM, model task dependency with probabilistic decomposition, meanwhile introducing IEB and PIP inadvertently. A promising direction is to conduct cascaded counterfactual regularization over these models to obtain unbiased estimation.

Another direction seeks to address the propensity score's fragility. The unbiasedness of IPS/DR methods heavily relies on the precision of propensity score (\ie CTR) estimation. However, obtaining accurate CTR estimation remains to be a thriving research topic within the recommender system community. A promising direction is to replace the weighting-based distribution alignment methods (\ie IPS/DR) with those based on adversarial training \cite{adda} or representation learning \cite{wda}.
\begin{acks}
We would like to extend our gratitude to the Ant Insurance Marketing Algorithm Group for sharing their experiences with us during the data preparation stage and the online deployment process.
\end{acks}

\bibliographystyle{ACM-Reference-Format}
\balance
\bibliography{sample-base}
\appendix

\section{Implementation Details}
% \textcolor{blue}{
In this section, we briefly describe the implementation procedure of ESCM$^2$. To construct ESCM$^2$-IPS:
\begin{enumerate}
    \item Construct a ESMM model with the CTR loss $\mathcal{L}_\mathrm{CTR}$ and CTCVR loss $\mathcal{L}_\mathrm{CTCVR}$ in the \emph{entire space} in Equation~\ref{eq:esmmLoss}. 
    \item Calculate the propensity score. In the CVR estimation task, a shortcut is to use the CTR estimate directly.
    \item Calculate the CVR estimation error $\mathcal{L}_\mathrm{CVR}$ with $\delta$ and propensity score in the \emph{clicked space}, following Equation~\ref{eq:ipsreg}.
    \item Calculate the learning objective of ESCM$^2$ following Equation~\ref{eq:escm}, and optimize it with stochastic gradient methods.
\end{enumerate}
% }
% \textcolor{blue}{
In step 2, small value of propensity scores (CTR estimates) leads to large estimation variance and numerical error. In practice, we set a threshold (\eg 0.1 in our setting) to clip the propensity score to a reasonable range following~\citet{treatment}. 
In step 3, the gradient from $\mathcal{L}_\mathrm{CVR}$ to propensity score (CTR estimate) ought to be truncated, otherwise $\mathcal{L}_\mathrm{CVR}$ would bias the CTR estimation.
In step 4, make every effort to ensure an accurate CTR estimate, as it has a significant impact on both the CVR and CTCVR estimates.
For example, set the weight $\lambda_\mathrm{c}$ to a small value (0.1-1 in our cases) as discussed in Figure~\ref{fig:lambda_c}, and enlarge the volume of CTR tower.
% }

% \textcolor{blue}{
The implementation of ESCM$^2$-DR follows the same procedure, while $\mathcal{L}_\mathrm{CVR}$ should be calculated with Equation~\ref{eq:drreg} instead. Alternative training is a useful trick to stabilize the training of DR regularizer, see the implementation for reference\textsuperscript{\ref{autodebias}}.
% }
\end{document}

%% file: 1_Introduction.tex
\section{Introduction}
\begin{figure}
\centering
\includegraphics[width=\linewidth]{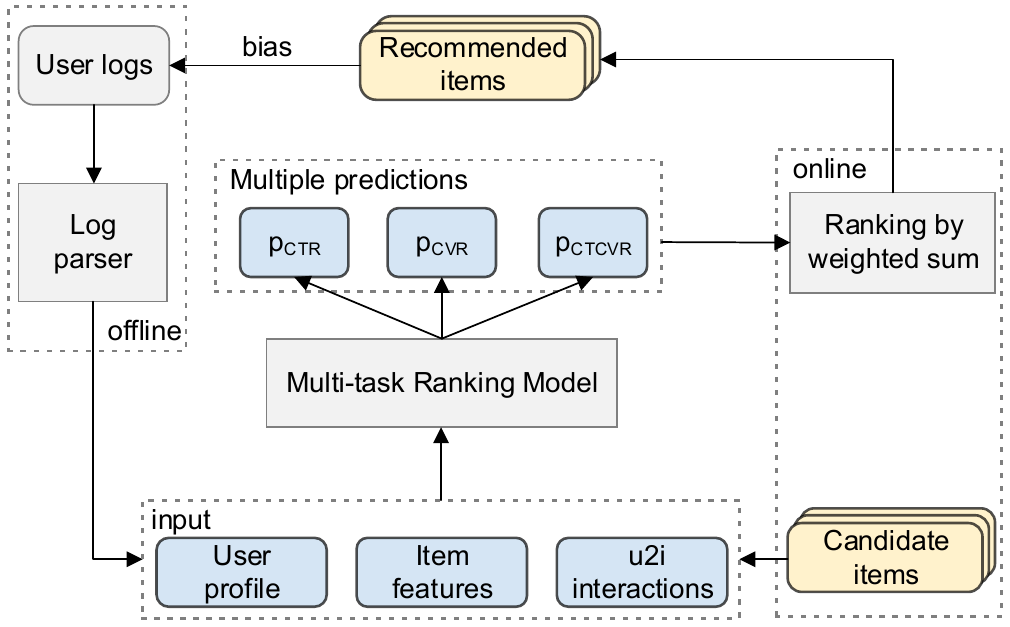}
\caption{A diagram of recommender system in e-commerce.\label{fig:recsys}}
\end{figure}

\noindent Recommender system aims to deliver valuable items from a large body of candidates to users \cite{deepfm,tianq}, which has been the main driving force of user growth in e-commerce
\cite{esmm}, social media \cite{social}, and advertising \cite{zhou2018deep}. \figref{fig:recsys} exhibits a two-stage pipeline to building recommender system in an industrial setting. During the offline stage, the ranking model is trained with user-profile features, item features and user-item-interaction features parsed from user logs. During the online stage, we rely on a number of ranking metrics, including but not limited to click-through rate (CTR), post-click conversion rate (CVR), and click-through\&conversion rate (CTCVR) to expose to the user the items that might capture his/her interest. Feedback from users is utilized during the offline training stage to optimize the performance of the recommender system. 

\begin{figure}
\centering
\includegraphics[width=\linewidth]{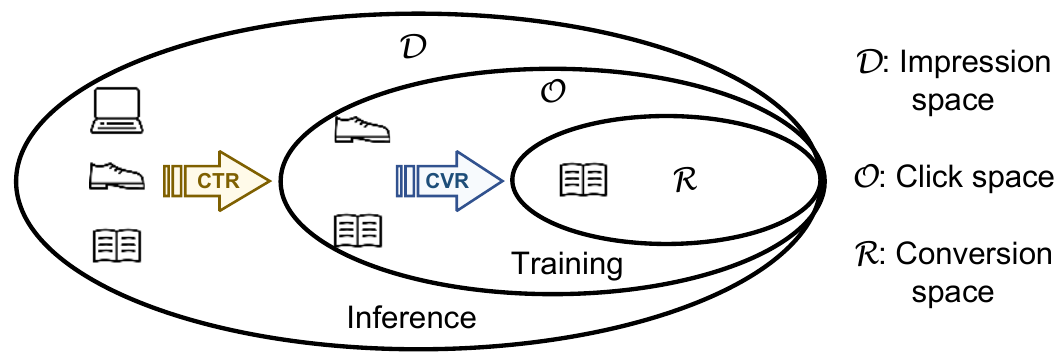}
\caption{Illustration of sample selection bias and data sparsity in CVR estimation task, where the training space only consists of clicked samples, while the inference space is the entire space for all impression samples.\label{fig:yangcong}}
\end{figure}

A typical user behavior path in e-commerce setting can be summarized as $impression\to click\to conversion$ \cite{esmm,esmm3}. As shown in \figref{fig:yangcong}, CVR denotes the transition probability from the click space to the conversion space. One naive approach to building effective CVR estimator is to train it over the click space since the conversion response is fully available in the click space.

Two critical issues that challenge na\"ive CVR estimators have been reported by \citeauthor{esmm} \cite{esmm} and \citeauthor{mtlips} \cite{mtlips}.
The first problem is \textbf{sample selection bias} due to the training space composing solely of clicked items. Specifically, items with lower CVR are less likely to be clicked, \ie to be included in the training space, and vice versa \cite{marlin2012collaborative}, which makes the training space Missing Not At Random. In other words, there is a distribution shift between the training space $\mathcal{O}$ and the inference space $\mathcal{D}$. Another problem stems from the \textbf{data sparsity} of clicked samples, where we have a CTR around 3.8\% on our production dataset and a 4\% on the Ali-CCP dataset. As na\"ive CVR models are trained over the click space, both problems severely hinder their generalization beyond click space.

\citeauthor{esmm} proposed the Entire Space Multi-task Model (ESMM) to address sample selection bias and data sparsity. Admittedly, it has mitigated data sparsity effectively via parameter sharing \cite{esmm,esm2,esmm3}; however, unbiasedness of its CVR estimate is still not guaranteed. Emerging empirical evidence \cite{mtlips} suggests that ESMM's CVR estimation is biased, raising concerns of this approach.

In this paper, we report two critical issues of ESMM in Section 3. 
\begin{itemize}
    \item \textbf{Inherent Estimation Bias (IEB):} We rigorously demonstrate that ESMM's CVR estimate is inherently higher than the ground truth even under very relaxed conditions.
    \item \textbf{Potential Independence Priority (PIP):} We demonstrate that ESMM's CTR and CVR estimates are prone to be conditionally independent, which is undesirable.
\end{itemize}

Leveraging causality methodology, we propose the Entire Space Counterfactual Multi-task Model (ESCM$^2$), a model that incorporates counterfactual risk minimizer (CRM), \ie the inverse propensity score (IPS) and the doubly robust (DR), to regularize ESMM's CVR estimation. 
As we will discuss in Section 4, the introduced CRM regularizer frees the CVR and the CTCVR estimates from the IEB and the PIP issues, respectively.

The contributions of this paper are summarized as follows.
\begin{itemize}
    \item This is the first work that rigorously demonstrates the inherent bias of ESMM's CVR estimates. Mathematical proofs and experiment results are provided to support this claim\footnote{\citeauthor{mtlips} \cite{mtlips} provided some numerical examples to showcase that ESMM's CVR estimates are bias, while the theoretical analysis was not rigorous. We provide a complete and easy-to-follow proof based on probabilistic tools in \thmref{thm:ieb}.}.
    \item We show that the ESMM's CTCVR estimates are subjected to potential independence priority. We have designed experiments to back up this claim.
    \item We propose ESCM$^2$, the first work that improves ESMM from a causal perspective. ESCM$^2$ effectively eliminates IEB and PIP in ESMM. Extensive experimental results and mathematical proofs are provided to verify our claims.
\end{itemize}

%% file: 2_preliminary.tex
\section{Preliminaries}
\subsection{Notations}
We use uppercase letter, \eg  $O$ to denote a random variable and lowercase letter, \eg $o$ to denote an associated specific value. Letters in calligraphic font, \eg $\mathcal{O}$ represent the sample space of the corresponding random variable, and $\mathbb{P}()$ represents the probability distribution of the random variable, \eg,  $\mathbb{P}(O)$.
\subsection{Problem Formulation}
Denote $\mathcal{U}=\{u_1, u_2,...,u_m\}$ as the set of $m$ users over the exposure space. Denote $\mathcal{I}=\{i_1, i_2,...,i_n\}$ as the set of $n$ items over the exposure space, and denote $\mathcal{D}=\mathcal{U}\times\mathcal{I}$ as the set of Cartesian product of the user set and the item set over the exposure space. Denote $\mathcal{O}$ as the click matrix where each entry $o_{u,i}\in\{0,1\}$ indicates whether clicking took place between user ${u}$ and item $i$, $\mathbf{R}\in\{0,1\}^{m\times n}$ as the observed conversion label where each entry $r_{u,i}\in\{0,1\}$ indicates whether conversion takes place between user $u$ and item $i$. 

If $\mathbf{R}$ is fully observed, the ideal loss function is formulated as:
\begin{equation}\label{eq:ideal}
    \mathcal{P}:=\mathbb{E}_{(u,i)\in\mathcal{D}}\left[\delta\left(r_{u,i}, \hat{r}_{u,i}\right)\right],
\end{equation}
where $\delta$ denotes the prediction error of $r_{u,i}$, $\hat{r}_{u,i}$ denotes the predicted $r_{u,i}$. For $\delta$, we use the binary cross entropy loss:
\begin{equation}
    \delta(r_{u,i}, \hat{r}_{u,i}):=-r_{u,i}log\hat{r}_{u,i}-(1-r_{u,i})log(1-\hat{r}_{u,i}).
\end{equation}

However, $r_{u,i}$ can only be observed for user-item pairs in the click space $\mathcal{O}$. As such, a na\"ive approach \cite{treatment,dr} estimates the ideal loss with expectation over $\mathcal{O}$:
\begin{equation}
\label{eq:naive}
    \mathcal{L}_\mathrm{naive}=\mathbb{E}_{(u,i)\in\mathcal{O}}(\delta_{u,i})=\frac{1}{|\mathcal{O}|}\sum_{(u,i)\in\mathcal{D}}(o_{u,i}\delta_{u,i}),
\end{equation}
where $|\mathcal{O}|=\sum_{(u,i)\in\mathcal{D}}(o_{u,i})$. It is widely adopted by many existing methods, but leads to biased estimation, \ie $\mathbb{E}_{\mathcal{O}}[\mathcal{L}_\mathrm{naive}]\neq\mathcal{P}$.
\subsection{Entire Space Multitask Model Approach}
The Entire Space Multitask Model Approach (ESMM) \cite{esmm} uses chain rule to obtain CVR estimate indirectly:
\begin{equation}
\label{eq:esmmDecom}
    \mathbb{P}(r_{u,i}=1\mid o_{u,i}=1)=\frac{\mathbb{P}(r_{u,i}=1,o_{u,i}=1)}{\mathbb{P}(o_{u,i}=1)}.
\end{equation}

Two towers are used to predict CTR (\ie$\mathbb{P}(o_{u,i}=1)$) and CVR (\ie$\mathbb{P}(r_{u,i}=1\mid o_{u,i}=1)$) in ESMM, respectively. The product of these two towers gives CTCVR estimation (\ie$\mathbb{P}(r_{u,i}=1,o_{u,i}=1)$). During the training stage, ESMM minimizes the empirical risk of CTR and CTCVR estimation over the entire impression space $\mathcal{D}$:
\begin{equation}
\begin{aligned}
\mathcal{L}_\mathrm{CTR} &= \mathbb{E}_{(u,i)\in\mathcal{D}}\left[\delta\left(o_{u,i},\hat{o}_{u,i}\right)\right]\\
\mathcal{L}_\mathrm{CTCVR} &= \mathbb{E}_{(u,i)\in\mathcal{D}}\left[\delta\left(o_{u,i}*r_{u,i},\hat{o}_{u,i}*\hat{r}_{u,i}\right)\right].
\end{aligned}
\label{eq:esmmLoss}
\end{equation}

During the inference stage, ESMM uses the output from the CVR tower as its predicted CVR. This approach circumvents the sample selection bias problem by not modeling CVR over the click space. However, as illustrated in Section 4.1, such approach yields inherent overestimation of CVR. To that end, we seek to develop an unbiased CVR estimator to address sample selection bias appropriately. 
\subsection{Propensity Score Based Approach}
The inverse propensity score (IPS) estimator \cite{treatment} weights
each error term ${\delta_{u, i}}$ with $1/q_{u,i}$, the inverse of the propensity score (CTR in our case), to align the error distribution over the click space with that over the exposure space. The adjusted error term is:
\begin{equation}
\label{eq:ips}
    \mathcal{L}_\mathrm{IPS}
    =\frac{1}{|\mathcal{D}|}\sum_{(u,i)\in\mathcal{D}}\frac{o_{u,i}\delta_{u,i}}{q_{u,i}}
    =\frac{1}{|\mathcal{D}|}\sum_{(u,i)\in\mathcal{D}}\frac{o_{u,i}\delta_{u,i}}{\hat{q}_{u,i}}.
\end{equation}

As the real value $q_{u,i}$ is always unavailable, an auxiliary classifier is introduced to estimate the propensity score $q_{u,i}$ with $\hat{q}_{u,i}$. The IPS estimator renders an unbiased estimate of the ideal loss function, \ie $\mathbb{E}_{\mathcal{O}}(\mathcal{L}_\mathrm{IPS})=\mathcal{P}$, given that the estimated $\hat{q}_{u,i}$ is accurate \cite{treatment}.

However, propensity score in the IPS estimator might suffer from severely high variance, for which the doubly robust (DR) estimator is introduced \cite{dr}. In particular, DR introduces imputed error $\hat{\delta}_{u,i}$ to model the prediction error for all events in $\mathcal{D}$, and corrects the error deviation $\hat{e}_{u,i}=\delta_{u,i}-\hat{\delta}_{u,i}$ for clicked events:
\begin{equation}
\label{eq:dr}
    \mathcal{L}_\mathrm{DR}=\frac{1}{|\mathcal{D}|} \sum_{(u, i) \in \mathcal{D}} \hat{\delta}_{u, i}+\frac{o_{u, i}\hat{e}_{u,i}}{\hat{q}_{u, i}},
\end{equation}
where $\hat{q}_{u,i}$ seeks to eliminate the MNAR effect for $\hat{e}_{u,i}$. Double robustness derives from the fact that unbiasedness is guaranteed as long as either the imputation error or the propensity score is accurate, but not necessarily both. The accuracy of $\hat{\delta}_{u,i}$ and $\hat{q}_{u,i}$ is usually ensured by auxiliary tasks.

%% file: 3_discussion.tex
\section{Discussion on ESMM}

\subsection{Is ESMM an Unbiased CVR Estimator?}

The existence of \textbf{inherent estimation bias (IEB)} in ESMM has been perceived by researchers \cite{mtlips}; however, to the best of our knowledge, a theoretical proof of its CVR estimation bias remains lacking. In this paper, IEB is formulated and proved in \thmref{thm:ieb}.  
\begin{thm}
\label{thm:ieb}
Let random variables $O,R,C$ be the indicator of click, post-click conversion and click \& conversion, and $o_{u,i}, r_{u,i}, c_{u,i}$ be the corresponding value of $O,R,C$ given user-item pairs, and $\hat{o}_{u,i}, \hat{r}_{u,i}, \hat{c}_{u,i}$ be the predicted value of $o_{u,i}, r_{u,i}, c_{u,i}$. 
The bias of ESMM's CVR estimate over the exposure space $\mathcal{D}$ is always larger than zero:
\begin{equation}
    Bias^\mathrm{ESMM}:=
    \mathbb{E}_{\mathcal{D}}\left[\hat{R}\right]-\mathbb{E}_{\mathcal{D}}\left[R\right]>0,
\end{equation}
given that users in click space are more likely to be converted \cite{treatment}, \ie
\begin{equation}
\label{eq:assumption}
    \mathbb{E}_{\mathcal{O}}\left[R\right]>\mathbb{E}_{\mathcal{D}}\left[R\right].
\end{equation}
\end{thm}
\begin{proof}
Following the loss function in \eqaref{eq:esmmLoss}, a well-trained ESMM model ensures:
\begin{equation}
\begin{aligned}
    \mathbb{E}_{\mathcal{D}}\left[O-\hat{O}\right]&=\int \left(o_{u,i}-\hat{o}_{u,i}\right) d(u,i)=0\\
    \mathbb{E}_{\mathcal{D}}\left[C-\hat{C}\right]&=\int \left(c_{u,i}-\hat{c}_{u,i}\right) d(u,i)=0.
\end{aligned}
\label{eq:esmm2}
\end{equation}

Noting that $\mathbb{E}_{\mathcal{D}}\left[R\right]$ and $\mathbb{E}_{\mathcal{D}}\left[\hat{R}\right]$ are the expectation of CVR ground truth and estimate, respectively. The CVR estimation bias:
\begin{equation}
\begin{aligned}
Bias^\mathrm{ESMM}&=\mathbb{E}_{\mathcal{D}}\left[\hat{R}\right]-\mathbb{E}_{\mathcal{D}}\left[R\right]\\
&\overset{(1)}{>}\mathbb{E}_{\mathcal{D}}\left[\hat{R}\right]-\mathbb{E}_{\mathcal{O}}\left[R\right]\\
&\overset{(2)}{=}\mathbb{E}_{\mathcal{D}}\left[\hat{R}\right]-\frac{\mathbb{E}_{\mathcal{D}}\left[C\right]}{\mathbb{E}_{\mathcal{D}}\left[O\right]}\\
&\overset{(3)}{=}\mathbb{E}_{\mathcal{D}}\left[\frac{\hat{C}}{\hat{O}}\right]-\frac{\mathbb{E}_{\mathcal{D}}\left[C\right]}{\mathbb{E}_{\mathcal{D}}\left[O\right]},\\
\end{aligned}
\label{eq:bias}
\end{equation}
where 
(1) comes from the assumption in \eqaref{eq:assumption}; 
(2) estimates the ground-truth CVR expectation over $\mathcal{D}$ following the label distribution regime in \figref{fig:yangcong}, 
(3) stems from the probability decomposition of ESMM in \eqaref{eq:esmmDecom}.

Let $\mathbb{P}(\hat{c}, \hat{o})$ be the joint probability of $\hat{C}=\hat{c}$ and $\hat{O}=\hat{o}$ over exposure space. The first term in \eqaref{eq:bias} is formulated as
\begin{equation}
\begin{aligned}
\mathbb{E}_{\mathcal{D}}\left[\frac{\hat{C}}{\hat{O}}\right] 
&\overset{(1)}{=} \int \frac{\hat{c}}{\hat{o}}\mathbb{P}(\hat{c}, \hat{o})d(\hat{c},\hat{o}) \\
&\overset{(2)}{=} \int \frac{\hat{c}}{\hat{o}}\mathbb{P}(\hat{c})\mathbb{P}(\hat{o})d(\hat{c},\hat{o}) \\
&\overset{(3)}{=} \int\hat{c}\mathbb{P}(\hat{c})d\hat{c}\int\frac{1}{\hat{o}}\mathbb{P}(\hat{o})d\hat{o} \\
&\overset{(4)}{=} \mathbb{E}_{\mathcal{D}}\left[\hat{C}\right]\mathbb{E}_{\mathcal{D}}\left[\frac{1}{\hat{O}}\right]\\
&\overset{(5)}{\geq} \frac{\mathbb{E}_{\mathcal{D}}\left[\hat{C}\right]}{\mathbb{E}_{\mathcal{D}}\left[\hat{O}\right]}
\overset{(6)}{=} \frac{\mathbb{E}_{\mathcal{D}}\left[C\right]}{\mathbb{E}_{\mathcal{D}}\left[O\right]}.
\end{aligned}
\label{eq:bias2}
\end{equation}
Below is some explanation for the derivation:
\begin{enumerate}[(1)]
	\item is the expectation of the random variable $\hat{C}/\hat{O}$.
	\item holds under the assumption $\hat{O}\upmodels\hat{C}$.
    \item breaks down the integrals of the product into the product of the integrals, and get the expectations in (4).
    \item[(5)] holds because we have Jensen's inequality $\mathbb{E}\left[f(X)\right]\geq f\left(\mathbb{E}\left[X\right]\right)$ for convex function $f(X)=1/X$. The equality holds only when the variance of $X$ equals to zero.
    \item[(6)] holds given a fully trained ESMM in \eqaref{eq:esmm2}.
 \end{enumerate}
 
As such, ESMM's CVR estimation is always larger than the ground-truth, \ie, $Bias^\mathrm{ESMM}>0$. The proof is completed.
\end{proof}
\thmref{thm:ieb} demonstrates the inherent bias of ESMM's CVR estimate. It inspires us to propose a CVR estimator over click space directly and address the IEB problem explicitly, rather than circumventing this critical problem via the entire-space modeling approach.
\subsection{Is ESMM a well-defined CTCVR Estimator?}
\begin{figure}
    % \addtocounter{figure}{-1}
    \centering
    \subfigure[ESMM Approach]{\includegraphics[width=0.28\linewidth,trim=0 0 -3 0,clip]{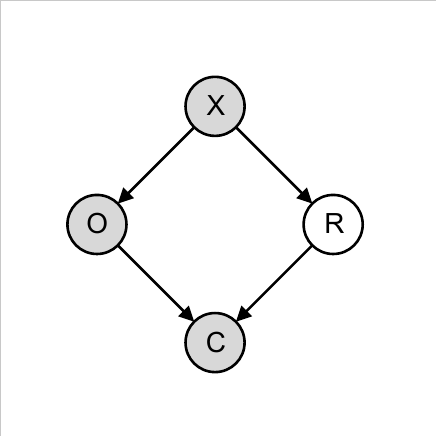}}\hspace{5mm}
    \subfigure[Na\"ive Approach]{\includegraphics[width=0.28\linewidth,trim=0 0 -3 0,clip]{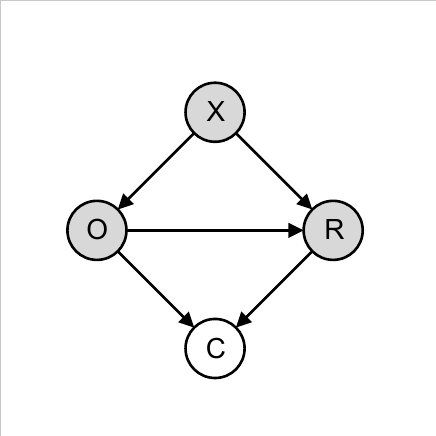}}\hspace{5mm}
    \subfigure[ESCM$^2$ Approach]{\includegraphics[width=0.28\linewidth,trim=0 0 -3 0,clip]{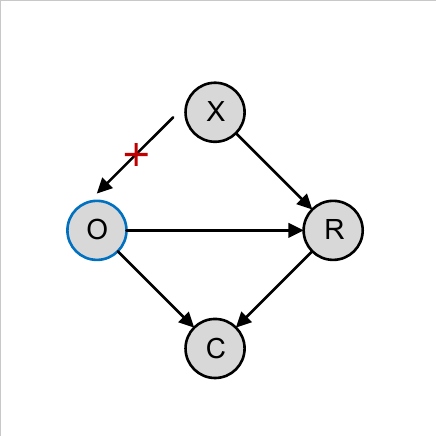}}
    \caption{Causal graphs depicting training of ESMM \cite{esmm}, Na\"ive \cite{mtlips} and ESCM$^2$, where X, O, R, C represent the user-item pair, click, conversion and click \& conversion, respectively. Hollow circles indicate latent variables, and shaded circles indicate observed variables. The intervention on O, represented as a blue circle in (c), eliminates the dependence between X and O, and thus removes the arrow X$\rightarrow$O.}
    \label{fig:causal}
\end{figure}
The causal graph of ESMM based on \eqaref{eq:esmmDecom} and \eqaref{eq:esmmLoss} is illustrated in \figref{fig:causal} (a). Specifically, ESMM uses two towers to predict CTR and CVR, and multiplies them to predict CTCVR:
\begin{equation}
\label{eq:ctcvr1}
    \mathbb{P}(o_{u,i}=1,r_{u,i}=1)=\mathbb{P}(o_{u,i}=1)*\mathbb{P}(r_{u,i}=1\mid o_{u,i}=1),
\end{equation}
where the CVR estimate depends on the click event. Specifically, conversion only happens after clicks. As such, there should be a causality link from $O$ to $R$ in the data generation process. However, in \figref{fig:causal} (a), this link is omitted by the entire space modeling strategy \cite{esmm}, as indicated by the missing arrow from $O\rightarrow R$.
The missing arrow poses the risk that, for CTCVR task, ESMM models CVR as $\mathbb{P}(r_{u,i}=1)$ as shown in \figref{fig:causal} (a), rather than the desired $\mathbb{P}(r_{u,i}=1\mid o_{u,i}=1)$. We formulate this risk as the potential independence priority problem since it replaces $\mathbb{P}(r_{u,i}=1\mid o_{u,i}=1)$ with $\mathbb{P}(r_{u,i}=1)$, and thus introduces false independent prior.

The na\"ive approach \cite{mtlips} in \figref{fig:causal} (b) trains the CVR model within the click space, and thus there is an arrow O$\rightarrow$R\footnote{This causal graph comes from Figure 1 in \cite{gu2021estimating} and Figure 1 in \cite{bareinboim2012controlling}.}.
However, the sample selection bias, indicated by the arrow X$\rightarrow$O brings sample selection bias \cite{esmm,mtlips}.
One approach to solve PIP without introducing selection bias is modeling CVR as $\mathbb{P}(r_{u,i}=1\mid \textit{do}(o_{u,i}=1))$ instead.
The "$do$" denotes the do-calculus, represented by the blue circle in the \figref{fig:causal} (c), truncates the dependency from $X$ to $O$.
For clicked samples, it is consistent with the conventional CVR problem, but for unclicked samples, it models a counterfactual problem: how likely is it that a user would be converted if he/she clicked the item?
Finally, the adjusted estimand of CTCVR is
\begin{equation}
    \mathbb{P}\left(o_{u,i}=1,r_{u,i}=1\right)=\mathbb{P}\left(o_{u,i}=1\right)*\mathbb{P}\left(r_{u,i}=1\mid do\left(o_{u,i}=1\right)\right),
\label{eq:ctcvr}
\end{equation}
according to the rule of do-calculus \cite{peal_2009}, which addresses PIP and selection bias simultaneously. The obtained CTCVR estimate is thus valid for the entire inference space.

%% file: 4_proposed.tex
\section{Proposed method}
\subsection{Counterfactual Risk Regularizer}
ESCM$^2$ leverages counterfactual risk minimizers based on inverse propensity score \cite{treatment,mtlips} and doubly robust \cite{dr,mrdr} to regularize ESMM. We further illustrate that these regularizers mitigate both IEB and PIP in ESMM.
\subsubsection{Inverse Propensity Score Regularizer}
\begin{figure*}
    \centering
    \includegraphics[width=0.9\textwidth]{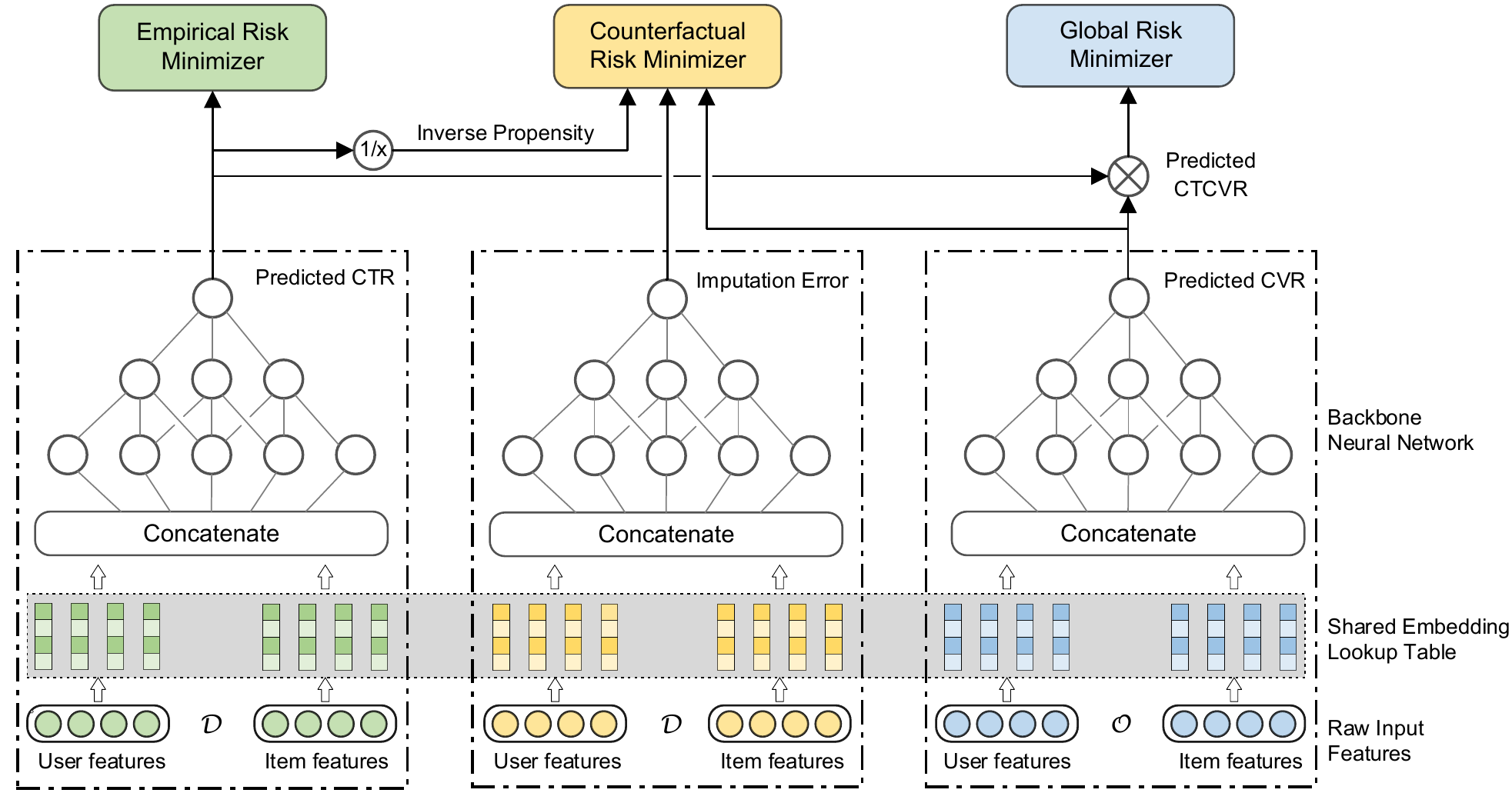}
    \caption{Architecture of ESCM$^2$-IPS and ESCM$^2$-DR, composed of the empirical risk minimizer for CTR estimation, the global risk minimizer for CTCVR estimation and the counterfactual risk minimizer for CVR estimation. All of the three optimizers share a common embedding lookup table. The ESCM$^2$-DR augments the ESCM$^2$-IPS with an imputation error tower.}
    \label{fig:modelStructure}
\end{figure*}

\ \\
Following \eqaref{eq:ips}, IPS regularizer inversely weights the CVR loss with $\hat{q}_{u,i}$, where $\hat{q}_{u,i}$ is usually modeled via a logistic CTR prediction model. As shown in \figref{fig:modelStructure}, \emph{we use the output of the CTR tower $\hat{o}_{u,i}$ in ESMM to estimate the propensity score ${q}_{u,i}$ in \eqaref{eq:ips}}. As such, the regularizer \cite{mtlips} can be formulated as:
\begin{equation}
\begin{aligned}
\label{eq:ipsreg}
\mathcal{R}_\mathrm{IPS}(\phi_{\mathrm{CTR}},\phi_{\mathrm{CVR}})
    &=\mathbb{E}_{(u, i) \in \mathcal{D}}\left[\frac{o_{u, i} \delta\left(r_{u, i}, \hat{r}_{u, i}\left(x_{u, i} ; \phi_{\mathrm{CVR}}\right)\right)}{\hat{o}_{u, i}\left(x_{u, i} ; \phi_{\mathrm{CTR}}\right)}\right]\\
    &=\frac{1}{|\mathcal{D}|} \sum_{(u, i) \in \mathcal{D}} \frac{o_{u, i} \delta\left(r_{u, i}, \hat{r}_{u, i}\left(x_{u, i}; \phi_{\mathrm{CVR}}\right)\right)}{\hat{o}_{u, i}\left(x_{u, i} ; \phi_{\mathrm{CTR}}\right)},
\end{aligned}
\end{equation}
where $x_{u,i}$ denotes the concatenated embedding given the user-item pair, and $\delta_{u,i}$ represents the estimation error of CVR. $\phi_{\mathrm{CTR}}$ and $\phi_{\mathrm{CVR}}$ are parameters of the CTR and the CVR tower in ESMM.

We extend the theorem of \citet{mtlips} to demonstrate that $\mathcal{R}_\mathrm{IPS}$ is equivalent to $\mathcal{P}$ over the exposure space, which gives us unbiased CVR estimation and hence mitigate the inherent estimation bias (IEB) of CVR estimation in ESMM.

\begin{thm}
\label{thm:anti-ieb}
Let $o_{u,i}$ be the indicator of whether user $u$ clicks on item ${i}$, $\hat{o}_{u,i}$ be the CTR estimate by ESMM with IPS regularizer. Given that the CTR is estimated accurately, \ie $\hat{o}_{u,i}=o_{u,i}$, the IPS-based counterfactual regularizer $\mathcal{R}_\mathrm{IPS}$ is equivalent to the ideal loss function in Equation~\ref{eq:ideal}, \ie $\mathcal{R}_\mathrm{IPS}=\mathcal{P}$, encouraging unbiased CVR estimation.
% \begin{equation}
%     \mathcal{R}_\mathrm{IPS}=\mathcal{P}.
% \end{equation}
% \begin{equation}
%     \mathbb{E}_{(u,i)\in\mathcal{D}}\left[\mathcal{R}_\mathrm{IPS}\right]-\mathcal{P}=0.
% \end{equation}
\end{thm}
% \begin{proof}
% \begin{equation}
%     \begin{aligned}
%     \mathbb{E}_{(u,i)\in\mathcal{D}}\left[\mathcal{R}_\mathrm{IPS}\right]
%     &=\frac{1}{|\mathcal{D}|} \sum_{(u, i) \in \mathcal{D}} \mathbb{E}_{\mathcal{D}}\left[\frac{o_{u, i} \delta\left(r_{u, i}, \hat{r}_{u, i}\right)}{\hat{q}_{u, i}}\right]\\
%     &=\frac{1}{|\mathcal{D}|} \sum_{(u, i) \in \mathcal{D}} \frac{q_{u, i} \delta\left(r_{u, i}, \hat{r}_{u, i}\right)}{\hat{q}_{u, i}}\\
%     &=\frac{1}{|\mathcal{D}|} \sum_{(u, i) \in \mathcal{D}} \delta\left(r_{u, i}, \hat{r}_{u, i}\right)=\mathcal{P}.
%     \end{aligned}
% \end{equation}
% \end{proof}
\begin{proof}
% \textcolor{blue}{
\begin{equation}
    \begin{aligned}
    \mathcal{R}_\mathrm{IPS} 
    &= \mathbb{E}_{\mathcal{D}}\left[\frac{o_{u, i} \delta\left(r_{u, i}, \hat{r}_{u, i}\right)}{\hat{o}_{u, i}}\right] \\
    &= \frac{|\mathcal{O}|}{|\mathcal{D}|}\mathbb{E}_{\mathcal{O}}\left[\frac{ \delta\left(r_{u, i}, \hat{r}_{u, i}\right)}{\hat{o}_{u, i}}\right] \\
    &= \frac{|\mathcal{O}|}{|\mathcal{D}|}\int\frac{ \delta\left(r_{u, i}, \hat{r}_{u, i}\right)}{\hat{o}_{u, i}} \mathbb{P}(u,i\mid O=1)\, d(u,i) \\
    &\overset{(1)}{=} \frac{|\mathcal{O}|}{|\mathcal{D}|}\int\frac{ \delta\left(r_{u, i}, \hat{r}_{u,i}\right)}{\mathbb{P}(O=1\mid u,i)} \mathbb{P}(u,i\mid O=1)\, d(u,i) \\
    &\overset{(2)}{=} \int\delta\left(r_{u, i}, \hat{r}_{u, i}\right) \mathbb{P}(u,i)\, d(u,i) = \mathcal{P}.
    \end{aligned}
\end{equation}
% }
where (1) holds because we have $\hat{o}_{u,i}=o_{u,i}$; (2) holds because we conduct the Bayes' theorem on $\mathbb{P}(u,i\mid O=1)$.
\end{proof}
Also, we show that $\mathcal{R}_\mathrm{IPS}$ models CVR as $\mathbb{P}\left(r_{u, i}=1 \mid d o\left(o_{u, i}=1\right)\right)$, as shown in \eqaref{eq:ctcvr}. It concurs with the causal dependency of the conversion event on the click event, and thus mitigates potential independence priority (PIP) of the CTCVR estimation in ESMM.
\begin{thm}
\label{thm:anti-pip}
Let $\mathbb{P}\left(r_{u, i}=1 \mid d o\left(o_{u, i}=1\right)\right)$ be the counterfactual conversion rate given the click event happens,  $\hat{r}^\mathrm{IPS}_{u,i}$ be the predicted CVR of ESMM with IPS regularizer. For all user-item pairs in the exposure space, the IPS regularizer encourages:
\begin{equation}
    \hat{r}_{u,i}^\mathrm{IPS}\rightarrow\mathbb{P}\left(r_{u, i}=1 \mid d o\left(o_{u, i}=1\right)\right).
\end{equation}
\end{thm}
\begin{proof}

Partition the user-item pairs into $K$ boxes $L_1$, $L_2$, ..., $L_K$ such that conditional exchangeability holds in each box $L_k$.
That is, within the group $L_k$, the observed distribution of click events $O$ is independent of the counterfactual error of CVR estimation:
\begin{equation}
    \label{eq:exchange}
    \{\delta^{(0)}_k, \delta^{(1)}_k\}\upmodels O_k,
\end{equation}
where $\delta^{(1)}_k$ is the counterfactual distribution of CVR estimation error if $o_{u, i} = 1$ for all $(u,i)\in L_k$, $\delta^{(0)}_k$ is the counterfactual distribution of error if $o_{u, i} = 0$ for all $(u,i)\in L_k$. 

Leveraging the law of iterated expectations:
\begin{equation}
    \mathcal{R}_{\mathrm{IPS}}
    =\mathbb{E}_{(u,i)\in\mathcal{D}}\left[\frac{o_{u, i}\delta_{u,i}}{\hat{q}_{u, i}}\right]
    =\mathbb{E}_{k}\left\{\mathbb{E}_{(u,i)\in L_k}\left[\frac{o_{u,i}}{\hat{q}_{u,i}} \delta_{u,i}\right]\right\},
\end{equation}
and within the group $L_k$, we have:
\begin{equation}
\begin{aligned}
    \mathbb{E}_{(u,i)\in L_k}\left[\frac{o_{u,i}}{\hat{q}_{u,i}} \delta_{u,i}\right]
    &\overset{(1)}{=}\mathbb{E}_{(u,i)\in L_k}\left[\frac{o_{u,i}}{\hat{q}_{u,i}} \delta_{u,i}^{(1)}\right]\\
    &\overset{(2)}{=}\mathbb{E}_{(u,i)\in L_k}\left[\frac{o_{u,i}}{\hat{q}_{u,i}}\right]\mathbb{E}_{(u,i)\in L_k}\left[\delta_{u,i}^{(1)}\right]\\
    &=\mathbb{E}_{(u,i)\in L_k}\left[\delta_{u,i}^{(1)}\right]\triangleq \Delta_k,
\end{aligned}
\end{equation}
% \textcolor{blue}{
where (1) and (2) holds because we have Equation~\ref{eq:exchange}
% }
;$\Delta_k$ denotes the expectation of $\delta_k^{(1)}$. Therefore, $\mathcal{R}_\mathrm{IPS}$ can be reformulated as:
\begin{equation}
    \mathcal{R}_{\mathrm{IPS}}
    =\mathbb{E}_{k}\left[\Delta_k\right].
\end{equation}

As such, minimizing $\mathcal{R}_\mathrm{IPS}$ is equivalent to minimizing $\Delta_k$ for the groups $k=1,2,...,K$. Particularly in the group $L_k$, minimizing $\Delta_k$ makes the $\hat{r}_{u,i}^\mathrm{IPS}$ converge to the ground-truth CVR given click happens for all samples in this group.
\end{proof}
So far we have proved that the IPS regularizer is able to alleviate IEB and PIP in ESMM. However, it still suffers from high variance problem \cite{mtlips,treatment} that makes the training process unstable. Therefore, we extend our scope to a doubly robust estimator,  which gives us lower variance and double robustness \cite{dr,mrdr}.
\subsubsection{Doubly Robust Regularizer}
\ \\
Following \eqaref{eq:dr}, the DR regularizer inversely weights the error deviation $\hat{e}_{u,i}=\delta_{u,i}-\hat{\delta}_{u,i}$, instead of $\delta_{u,i}$, with $\hat{o}_{u,i}$. As shown in \figref{fig:modelStructure}, we utilize the output of the CTR tower $\hat{o}_{u,i}$ in ESMM to estimate the propensity of click behavior, and the output of imputation tower $\hat{\delta}_{u,i}$ to predict the estimation error of CVR $\delta_{u,i}$. As such, the DR regularizer can be formulated as:
\begin{equation}
\label{eq:drreg}
\begin{aligned}
&\mathcal{R}_\mathrm{DR}^\mathrm{err}(\phi_{\mathrm{CTR}},\phi_{\mathrm{CVR}},\phi_{\mathrm{IMP}})\\
    &=\mathbb{E}_{(u, i) \in \mathcal{D}}\left[\hat{\delta}_{u,i}\left(x_{u,i};\phi_\mathrm{IMP}\right)+\frac{o_{u, i}\hat{e}_{u,i}\left(x_{u,i};\phi_\mathrm{CVR},\phi_\mathrm{IMP}\right)}{\hat{o}_{u, i}\left(x_{u, i} ; \phi_{\mathrm{CTR}}\right)}\right],
\end{aligned}
\end{equation}
where $x_{u,i}$ denotes the concatenated embedding given the user-item pair, $\delta_{u,i}$ the estimation error of CVR, $\phi_{\mathrm{CTR}}$ and $\phi_{\mathrm{CVR}}$ the parameters of the CTR and the CVR tower in ESMM, and $\phi_{\mathrm{IMP}}$ the parameter of the imputation tower.

The double robustness derives from the fact that the estimator is unbiased if either $\hat{\delta}_{u,i}$ or $\hat{o}_{u,i}$ is accurate. The accuracy of $\hat{o}_{u,i}$ is guaranteed by a properly trained CTR estimator. To ensure the accuracy of $\hat{\delta}_{u,i}$, an auxiliary imputation loss is introduced:
\begin{equation}
\begin{aligned}
&\mathcal{R}_\mathrm{DR}^\mathrm{imp}(\phi_{\mathrm{CTR}},\phi_{\mathrm{CVR}},\phi_{\mathrm{IMP}})\\
&=\mathbb{E}_{(u, i) \in \mathcal{D}}\left[\frac{o_{u,i}\hat{e}^2_{u,i}(x_{u,i};\phi_\mathrm{CVR},\phi_\mathrm{IMP})}{\hat{o}_{u,i}\left(x_{u, i} ; \phi_{\mathrm{CTR}}\right)}\right],
\end{aligned}
\end{equation}
and the doubly robust regularizer is formulated as:
\begin{equation}
\begin{aligned}
&\mathcal{R}_\mathrm{DR}(\phi_{\mathrm{CTR}},\phi_{\mathrm{CVR}},\phi_{\mathrm{IMP}})\\
&=\mathcal{R}_\mathrm{DR}^\mathrm{err}(\phi_{\mathrm{CTR}},\phi_{\mathrm{CVR}},\phi_{\mathrm{IMP}})+\mathcal{R}_\mathrm{DR}^\mathrm{imp}(\phi_{\mathrm{CTR}},\phi_{\mathrm{CVR}},\phi_{\mathrm{IMP}}).
\end{aligned}
\end{equation}
The DR regularizer also mitigates IEB and PIP in ESMM, since Theorems 2-3 also hold for the DR estimator. This proof is simple: we just need to replace the CVR estimation error $\delta$ in IPS with the error deviation $e$. We omit the exact proof for brevity.

\subsection{Architecture of ESCM$^2$}
As shown in \figref{fig:modelStructure}, we employ the multi-task learning technique to train ESCM$^2$, which is effective in alleviating data sparsity \cite{esmm,esm2,esmm3}. Specifically, the learning process is composed of three tasks: the empirical risk minimizer for CTR estimation, the global risk minimizer for CTCVR estimation, and the counterfactual risk minimizer for CVR estimation. 
The empirical risk minimizer aims to minimize the empirical risk of CTR estimation over the entire impression space $\mathcal{D}$:
\begin{equation}
    \mathcal{L}_\mathrm{CTR}(\phi_\mathrm{CTR})=\mathbb{E}_{(u, i) \in \mathcal{D}}\left[\delta\left(o_{u, i}, \hat{o}_{u, i}\left(x_{u,i};\phi_\mathrm{CTR}\right)\right)\right],
\end{equation}
where $\delta$ is binary cross-entropy in our case. Afterwards, the global risk minimizer optimizes the risk of CTCVR estimation over $\mathcal{D}$:
\begin{equation}
\begin{aligned}
&\mathcal{L}_\mathrm{CTCVR}(\phi_\mathrm{CTR}, \phi_\mathrm{CVR})\\
&=\mathbb{E}_{(u, i) \in \mathcal{D}}\left[\delta\left(o_{u, i}*r_{u, i}, \hat{o}_{u, i}\left(x_{u,i};\phi_\mathrm{CTR}\right)*\hat{r}_{u, i}\left(x_{u,i};\phi_\mathrm{CVR}\right)\right)\right].
\end{aligned}
\end{equation}

However, this approach suffers from inherent estimation bias (IEB) and potential independent priority (PIP). As such, we introduce the counterfactual risk minimizers in section 4.1 and derive the final learning objective of ESCM$^2$:
\begin{equation}
\label{eq:escm}
    \mathcal{L}_\mathrm{ESCM^2}:=\mathcal{L}_\mathrm{CTR}+\lambda_\mathrm{c}\mathcal{L}_\mathrm{CVR}+\lambda_\mathrm{g}\mathcal{L}_\mathrm{CTCVR},
\end{equation}
where $\lambda_\mathrm{c}$ and $\lambda_\mathrm{g}$ control the weights of counterfactual risk and global risk. Based on the formulation of $\mathcal{L}_\mathrm{CVR}$, we devise two specific implementations of ESCM$^2$, \ie ESCM$^2$-IPS and ESCM$^2$-DR:
\begin{itemize}
    \item ESCM$^2$-IPS: It computes the CVR risk as $\mathcal{L}_\mathrm{CVR}:= \mathcal{R}_\mathrm{IPS}$ in \eqaref{eq:ipsreg}. Specifically, it calculates the empirical risk of its CVR estimation over the click space, and inversely weights the risk with the output from its CTR tower. 
    \item ESCM$^2$-DR: It augments ESCM$^2$-IPS with imputation tower, and models the CVR risk with $\mathcal{L}_\mathrm{CVR}:=\mathcal{R}_\mathrm{DR}$ in \eqaref{eq:drreg}. 
\end{itemize}

As shown in \figref{fig:modelStructure}, the architecture of ESCM$^2$ exploits the sequential pattern of user actions, \ie $impression\rightarrow click\rightarrow conversion$ in e-commerce scenario. The amount of positive feedback for CTR task is significantly greater than that of CVR and CTCVR tasks by 1-2 order of magnitudes. Therefore, we share the feature representations learned in the CTR task with the CVR and the CTCVR tasks through a common look-up embedding table, which is effective in alleviating data sparsity.

%% file: 5_experiment.tex
\section{Experiments}
We conduct experiments to evaluate the performance of ESCM$^2$ and answer the following research questions:
\\\textbf{RQ1:} Does ESCM$^2$ outperform SOTA CVR and CTCVR estimators?
\\\textbf{RQ2:} Does ESMM inject bias on CVR estimation? To what extent does ESCM$^2$ reduce bias?
\\\textbf{RQ3:} Does ESMM inject independence prior when estimating CTCVR? To what extent does ESCM$^2$ alleviate this problem?
\\\textbf{RQ4:} How do critical components (\ie $\lambda_c,\lambda_g$) affect the performance of ESCM$^2$? Is ESCM$^2$ sensitive to these components?
\subsection{Experimental Setup}
\subsubsection{Datasets}
\begin{itemize}
    \item \textbf{Industrial dataset:}\footnote{The desensitized and encrypted data set does not contain any Personal Identifiable Information (PII). Adequate data protection was carried out during experiment to prevent the risk of data copy leakage, and the data set was destroyed after the experiment. The data set does not represent any business situation, only used for academic research} The industrial dataset comes from 90-day offline log of our business's recommender system, divided for training, validation and test in a chronological order. Afterwards, we downsample the negative samples of the training set to keep the ratio of exposure:click:conversion to be 100:10:1, approximately.
    \item \textbf{Public dataset:}\footnote{https://tianchi.aliyun.com/datalab/dataSet.html?dataId=408} The public dataset Ali-CCP (Alibaba Click and Conversion Prediction) is used to benchmark the performance of relevant methods for reproducibility consideration. Following \cite{seq2021}, all single-valued categorical features are fed into the models, and we randomly held out 10\% of the training dataset as our validation dataset. 
\end{itemize}
\begin{table}[]
    \caption{Dataset description.\label{tab:datasets}}
\resizebox{\linewidth}{!}
{
    \centering
    \begin{tabular}{c|cccccc}
    \hline\hline
         Dataset   &   \# User   &    \# Train  &   \# Valid   &    \# Test   & \# Click & \# Conversion \\ \hline
         Industrial  &  37.73M   &   61.58M      &   0.39M      & 24.28M
         & 3.73M & 0.32M
         \\
         Ali-CCP & 0.25M & 33.12M & 3.67M & 37.64M & 1.42M & 7.92K
         \\
          
         \hline\hline
    \end{tabular}
    }
    
\end{table}
\subsubsection{Competitors}
\ \\
Multi-task learning (MTL) has been shown to significantly improve recommender system's performance in many previous works \cite{mtlips,esmm3}. For fairness consideration, methods based on single-task learning (\eg the vanilla Inverse Propensity Score \cite{treatment} and Doubly Robust \cite{dr} approach) are not considered. There are two common methods which co-learn CTR and CVR simultaneously:
\begin{itemize}
    \item \textbf{Naïve}\footnote{https://github.com/PaddlePaddle/PaddleRec/tree/master/models/multitask\label{mtlmodel}} \cite{mmoe}: It directly models CVR as per \eqaref{eq:naive}, and shares embeddings across CTR and CVR tasks.
    \item \textbf{MTL-IMP} \cite{esmm}: It is implemented in the same way as the Naïve method, except that the unclicked samples are used directly as negative samples for CVR task.
\end{itemize}
The methods above have been known to induce selection bias for CVR estimation \cite{esmm}, and hence deteriorates our estimation for CTCVR. It is therefore necessary to include debiased approaches as our benchmark for fairness consideration.
\begin{itemize}
    \item \textbf{ESMM}\textsuperscript{\ref{mtlmodel}} \cite{esmm,esm2}: It leverages the multi-task learning to avoid the selection bias of CVR in an heuristic way.
    \item \textbf{MTL-EIB} \cite{eib}: The Error Imputation based approach leverages the imputed error for unclicked events and the prediction error for clicked events to get unbiased pcvr estimation.
    \item \textbf{MTL-IPS}\footnote{The MTL implementation is not available. See implementations for IPS and DR at https://github.com/DongHande/AutoDebias/tree/main/baselines for reference.\label{autodebias}} \cite{mtlips}: It implements the IPS estimator \cite{treatment} via a multi-task learning approach, which is theoretically unbiased.
    \item \textbf{MTL-DR}\textsuperscript{\ref {autodebias}} \cite{mtlips}: It is a multi-task learning version of DR estimator \cite{dr}, which is theoretically unbiased and more robust.
\end{itemize}

\subsubsection{Training Protocol}
\ \\
MMoE has been selected as the feature extractor for ESCM$^2$ and its competitors. It would also make sense to replace it with more powerful models such as AITM \cite{seq2021} and GemNN \cite{gemnn}. For fairness consideration, all models are trained for 300k iterations with Adam \cite{adam} optimizer and the same set of hyperparameters to make results comparable. We set the learning rate to be 1e$^{-4}$, and the weight decay to be 1e$^{-3}$. Other hyperparameters of the optimizer follow the literature \cite{adam}. Following \cite{seq2021}, the embedding dimension is set to 5. The objective weights $\lambda_\mathrm{g}$ and $\lambda_\mathrm{c}$ 
are set to 1 and 0.1, respectively, based on the parameter study in Section 5.5. We checkpoint the performance over the validation set every 1k iterations and export the best model to evaluate its performance on the test set. 
\subsubsection{Evaluation Protocol}
\ \\
For the offline experiments, following existing works, AUC (Area Under ROC) is primarily used as the main ranking metric to gauge performance. However, AUC only evaluates the average ranking performance of the model at all thresholds. To better understand how each model performs, we also report Kolmogorov-Smirnov (KS) score at the best threshold on ROC-curve and Recall and F1 score at the best threshold on PR-curve, respectively.
\begin{table*}[]
\caption{Offline performance (mean±std) on the CVR estimation task. Underlined results indicate the best baselines over each metric. "*" marks the methods that improve the best baselines significantly at p-value < 0.01 over paired samples t-test.}\label{tab:cvr}
\resizebox{\linewidth}{!}{
\begin{tabular}{c|ccccccccc}
\hline\hline
Dataset    & \multicolumn{4}{c}{Industrial Dataset} &  & \multicolumn{4}{c}{Ali-CCP Dataset} \\ \cline{2-5} \cline{7-10} 
Model      & AUC      & KS      & F1      & Recall  &  & AUC     & KS      & F1     & Recall \\ \hline
Naïve      & 0.7515±0.0164   & 0.3872±0.0043  & 0.3344±0.0052  & 0.5789±0.0071  &  & 0.5987±0.0139   & 0.1123±0.0056   & 0.0991±0.0053  & 0.2854±0.0050  \\
ESMM       & 0.7547±0.0183   & 0.3856±0.0051  & 0.6330±0.0074  & 0.5742±0.0055  &  & 0.6071±0.0133   & {\ul 0.1267±0.0043}   &  0.1157±0.0084  & 0.2968±0.0036  \\
MTL-EIB        & 0.7272±0.0140   & 0.3371±0.0051  & 0.5808±0.0048  & 0.5121±0.0072  &  & 0.5603±0.0135   & 0.0717±0.0057   & 0.0825±0.0051  & 0.2372±0.0043  \\
MTL-IMP & 0.7563±0.0114   & 0.3974±0.0047  & 0.1272±0.0040  & {\ul 0.5841±0.0092}  &  & {\ul 0.6114±0.0137}   & 0.1163±0.0043   & 0.1135±0.0055  & 0.2962±0.0056  \\
MTL-IPS    & {\ul 0.7586±0.0112}   & 0.3960±0.0048  & {\ul 0.6810±0.0042}  & 0.5651±0.0068  &  & 0.6091±0.0123   & 0.1177±0.0063   & 0.0941±0.2163  & {\ul 0.2975±0.0070}  \\
MTL-DR     & 0.7579±0.0135   & {\ul 0.4016±0.0046}  & 0.6804±0.0042  & 0.5137±0.0081  &  & 0.6065±0.0172   & 0.1255±0.0141  & {\ul 0.1159±0.0084}  & 0.2953±0.0178  \\
\hline
ESCM$^2$-IPS   & \textbf{0.7730±0.0150}   & \textbf{0.4144±0.0051$^{*}$}  & \textbf{0.7161±0.0089$^{*}$}  & 0.5932±0.0094  &  & \textbf{0.6163±0.0151}   & 0.1312±0.0060   & 0.1180±0.0047  & 0.3061±0.0059  \\
ESCM$^2$-DR    & 0.7679±0.0113   & 0.4119±0.0050  & 0.6884±0.0052  & \textbf{0.5986±0.0068$^{*}$}  &  & 0.6142±0.0133   & \textbf{0.1393±0.0042$^{*}$}   & \textbf{0.1315±0.0053$^{*}$}  & \textbf{0.3095±0.0054$^{*}$}  \\ \hline\hline
\end{tabular}
}
\end{table*}

\begin{table}[]
\caption{Offline performance on the CTCVR estimation task. “*” marks the methods that improve the best baselines significantly at p-value < 0.01 over paired samples t-test.}\label{tab:ctcvr}
\resizebox{\linewidth}{!}{
\begin{tabular}{c|ccccccccc}
\hline\hline
Dataset    & \multicolumn{4}{c}{Industrial Dataset}                                                                 &  & \multicolumn{4}{c}{Ali-CCP Dataset}                                                                    \\ \cline{2-5} \cline{7-10} 
Model      & \multicolumn{1}{c}{AUC} & \multicolumn{1}{c}{KS} & \multicolumn{1}{c}{F1} & \multicolumn{1}{c}{Recall} &  & \multicolumn{1}{c}{AUC} & \multicolumn{1}{c}{KS} & \multicolumn{1}{c}{F1} & \multicolumn{1}{c}{Recall} \\ \hline
Naïve      & 0.7954                  & 0.4631                 & 1.0048                 & 0.6602                     &  & 0.6003                  & 0.1192                 & 0.0978                 & 0.2921                     \\
ESMM       & {\ul 0.8153}            & 0.4827                 & 1.1062                 & {\ul 0.6819}               &  & 0.6081                  & 0.1292                 & {\ul 0.1139}           & {\ul 0.3027}               \\
MTL-EIB        & 0.7912                  & 0.4220                 & 0.8458                 & 0.5975                     &  & 0.5699                  & 0.0697                 & 0.0959                 & 0.2542                     \\
MTL-IMP & 0.7752                  & 0.4126                 & {\ul \textbf{1.3393}}  & 0.5880                     &  & 0.6087                  & 0.1264                 & 0.1110                 & 0.2973                     \\
MTL-IPS    & 0.8044                  & 0.4840                 & 1.0653                 & 0.6716                     &  & {\ul 0.6138}            & 0.1302                 & 0.1044                 & 0.2911                     \\
MTL-DR     & 0.8106                  & {\ul 0.4844}           & 1.1707                 & 0.6684                     &  & 0.6130                  & {\ul 0.1360}           & 0.1096                 & 0.2980                     \\ \hline
ESCM$^2$-IPS   & 0.8198                  & 0.4991                 & 1.1753                 & 0.6804                     &  & 0.6189                  & 0.1436                 & \textbf{0.1207$^{*}$}        & \textbf{0.3184$^{*}$}            \\
ESCM$^2$-DR    & \textbf{0.8265}         & \textbf{0.5134$^{*}$}        & 1.2842                 & \textbf{0.7013$^{*}$}            &  & \textbf{0.6245}         & \textbf{0.1494}        & 0.1180                 & 0.3117                     \\ \hline\hline
\end{tabular}
}
\end{table}
\subsection{Performance Comparison}
\subsubsection{Offline results}\ \\
We compare the offline performance of ESCM$^2$ and its competitors on both the industrial dataset and the public dataset, with mean and standard deviation reported over ten runs with different random seeds. We summarize the performance of the CVR estimation in \tabref{tab:cvr}, and get the following observations\footnote{To make the results more distinguishable, we report the F1 score in percentages.\label{ft:result}}:
\begin{itemize}
    \item Biased estimators achieve competitive performance on the CVR estimation task. Specifically, ESMM achieves an AUC of 0.754 on the industrial dataset, while the Naïve and the MTL-IMP estimators achieve AUCs of 0.751 and 0.756, respectively. Available empirical evidence \cite{esmm,mtlips} only demonstrates that ESMM outperforms biased estimators on single-task learning. Therefore, ESMM's better performance in our case might be attributed to its multi-task learning paradigm.
    \item Unbiased baseline estimators at large generally outperform the biased ones across these two datasets. For example, MTL-IPS achieves best performance over two metrics on the industrial dataset, which improves the KS and AUC of ESMM by 1.04\% and 0.39\%, respectively. It is therefore promising to regularize ESMM to get unbiased CVR estimation and better ranking performance. 
    \item The proposed ESCM$^2$ achieves significant improvement compared with various state-of-the-art baselines. Combined with aforementioned comparisons, We attribute its performance to unbiasedness of its estimation due to the counterfactual risk minimizers and the efficiency of the entire-space methodology without the IEB issue.
\end{itemize}

CTCVR is usually used as the primary ranking metric in industrial scenarios due to its empirical effectiveness. By eliminating both the IEB and the PIP issues, we expect better CTCVR estimation from ESCM$^2$ and hence better business metrics. We evaluate the performance of CTCVR estimation in \tabref{tab:ctcvr} and obtain the following observations\textsuperscript{\ref{ft:result}}:
\begin{itemize}
    \item Across all multi-task learning baselines, ESMM showcases competitive performance in the CTCVR estimation task. Specifically, ESMM achieves the highest AUC and Recall on the industrial dataset and the highest F1 and Recall on the Ali-CCP dataset. We attribute this to the fact that ESMM incorporates CTCVR explicitly into its learning objective. Benefiting from its superiority on CTCVR estimation, ESMM is widely deployed in numerous business scenarios.
    \item Although less significantly than on the CVR estimation task, ESCM$^2$ still beats competitors over most metrics. Specifically, ESCM$^2$-IPS achieves the best performance on F1 and Recall on Ali-CCP and ESCM$^2$-DR achieves the best performance on AUC and KS on the industrial dataset. We attribute ESCM$^2$'s better performance to the fact that it incorporates CTCVR into its learning objective directly and alleviates IEB and PIP through counterfactual risk regularization.
\end{itemize}

\subsubsection{Online A/B results}\ \\
\begin{table}
\caption{Results of Online A/B test}\label{tab:online}
\resizebox{\linewidth}{!}
{
\begin{tabular}{ccccccc}
\hline\hline
Metrics       & Day1     & Day2    & Day3    & Day4     & Day5    & Day6      \\
\hline
\# Order    & -9.76\%           & -1.85\% & -1.43\%  & \textbf{+9.07\%}  & \textbf{+0.73\%}           & \textbf{+6.26\%} \\
\# Premium  & \textbf{+64.53\%}  & \textbf{+37.47\%} & \textbf{+22.09\%} & -12.49\%           & \textbf{+4.26\%}  & \textbf{+11.10\%} \\
UV-CVR      & \textbf{+7.25\%}   & -1.66\%           & \textbf{+9.39\%}  & \textbf{+8.58\%}   & \textbf{+2.51\%}  & \textbf{+8.62\%}  \\
UV-CTCVR    & \textbf{+0.20\%}   & -3.50\%          & \textbf{+2.50\%}  & \textbf{+9.48\%}   & \textbf{+2.75\%}  & \textbf{+6.64\%}  \\
\hline\hline
\end{tabular}
}
\end{table}
We conducted online experiments to further showcase the superiority of ESCM$^2$ over ESMM.
Specifically, we implemented ESMM and ESCM$^2$ with our internal C++ based deep learning framework, where ESCM$^2$ was built with the IPS regularizer for its competitive offline performance and training efficiency. We then randomly assigned users into buckets, and observed each model's performance in respective bucket via our A/B testing platform. Live experiments were conducted in three large-scale scenarios. We are mainly interested in the models' performance on UV-CVR, UV-CTCVR and two additional business metrics: order quantity and actual premium.

\textbf{Scenario 1}: 
This experiment was deployed in one of Ant insurance's scenarios for 6 days, covering around 2.2 million unique visitors (UVs) and 3.1 million page views (PVs). Overall, ESCM$^2$ improved order quantity by 2.84\% and premium by 10.85\%. We also observed a significant 5.64\% and 3.92\% increase over UV-CVR and UV-CTCVR, respectively. Daily results are reported in \tabref{tab:online}, with ESCM$^2$ consistently outperforming ESMM on most metrics. 

\textbf{Scenario 2}: This experiment was deployed in the same but newly revamped Ant insurance's scenarios for six days, covering around 3.4 million UVs and 4.9 million PVs, respectively. Overall, ESCM$^2$ increased the order quantity by 4.26\% and the premium by 3.88\%. UV-CVR and UV-CTCVR also increased by 0.43\% and 1.75\%, respectively.

\textbf{Scenario 3}: This experiment was deployed in our Wufu campaign, covering around 125 thousand UVs and 136 thousand PVs for four days. Overall, ESCM$^2$ significantly increased order quantity by 40.55\% and premium by 12.69\%.

\subsection{Discussion on Inherent Estimation Bias}
We provide experimental results to support our claim that ESMM's CVR estimation suffers from IEB. Specifically the "Label" field in \tabref{tab:cvrbias} records the ground-truth conversion rate within the click space, which can be reasonably considered as an upper bound for the expected ground-truth CVR over the entire exposure space. Afterwards, we compare it with the mean of different estimators' CVR estimates over the exposure space.

The first observation from \tabref{tab:cvrbias} is that the ESMM's CVR estimates are consistently greater than the ground truth. For example, on the Ali-CCP trainset, the ground-truth CVR expectation does not exceed 0.0056, while ESMM reaches an expectation of 0.01. On the test set, ESMM overestimates CVR by 0.0057 on average. It concurs with our claim in \thmref{thm:ieb}.

The second observation is that counterfactual regularization significantly alleviates IEB, which supports our claims in Section 4.1. Specifically, on the industrial dataset, ESCM$^2$-IPS reduces the error of expectation by 0.031 on the training set and 0.035 on the test set; ESCM$^2$-DR significantly reduces the error by 0.037 and 0.045 on the training and test sets, respectively.
\begin{table}[]
\caption{Expectation of CVR estimates, where "Label" indicates the expectation of observed conversion labels.}
\resizebox{\linewidth}{!}
{
\begin{tabular}{cccccc}
\hline\hline
Dataset & Subset & Label & ESMM & ESCM$^2$-IPS & ESCM$^2$-DR \\
\hline
Ali-CCP  & Train   & 0.0055      & 0.0101±0.0011     &  0.0059±0.0005   &  0.0076±0.0018  \\
Ali-CCP  & Test    & 0.0056      & 0.0113±0.0008     &  0.0060±0.0009   &  0.0059±0.0009  \\
Industry & Train   & 0.0953      & 0.1588±0.0107     &  0.1277±0.0053   &  0.1216±0.0093  \\
Industry & Test    & 0.0407      & 0.1643±0.0095     &  0.1290±0.0092   &  0.1188±0.0022  \\
\hline\hline
\end{tabular}
}
\label{tab:cvrbias}
\end{table}
\subsection{Discussion on Potential Independence Priority}
PIP stems from the model failing to capture the intrinsic causality from clicking to conversion, as indicated by the missing edge O$\rightarrow$R in \figref{fig:causal} (a). We identify the causality strength from click to conversion to showcase the existence of PIP for ESMM's CTCVR estimation. 
Due to the presence of confounder in \figref{fig:causal}, causality cannot be quantified using statistical estimators, such as the Pearson's correlation coefficient. Therefore, we preprocess data with propensity score matching (PSM) to eliminate confounding effect, following the settings in the tutorial by \citeauthor{psm} \cite{psm}. Specifically, we consider the CTR and CVR estimates as the propensity scores and outcomes, respectively. We then divide samples into two groups according to the click indicator $O$. Each clicked sample is matched to some unclicked samples with similar propensity scores. 
\begin{figure}
    \centering
    \includegraphics[width=\linewidth,trim=20 10 20 10]{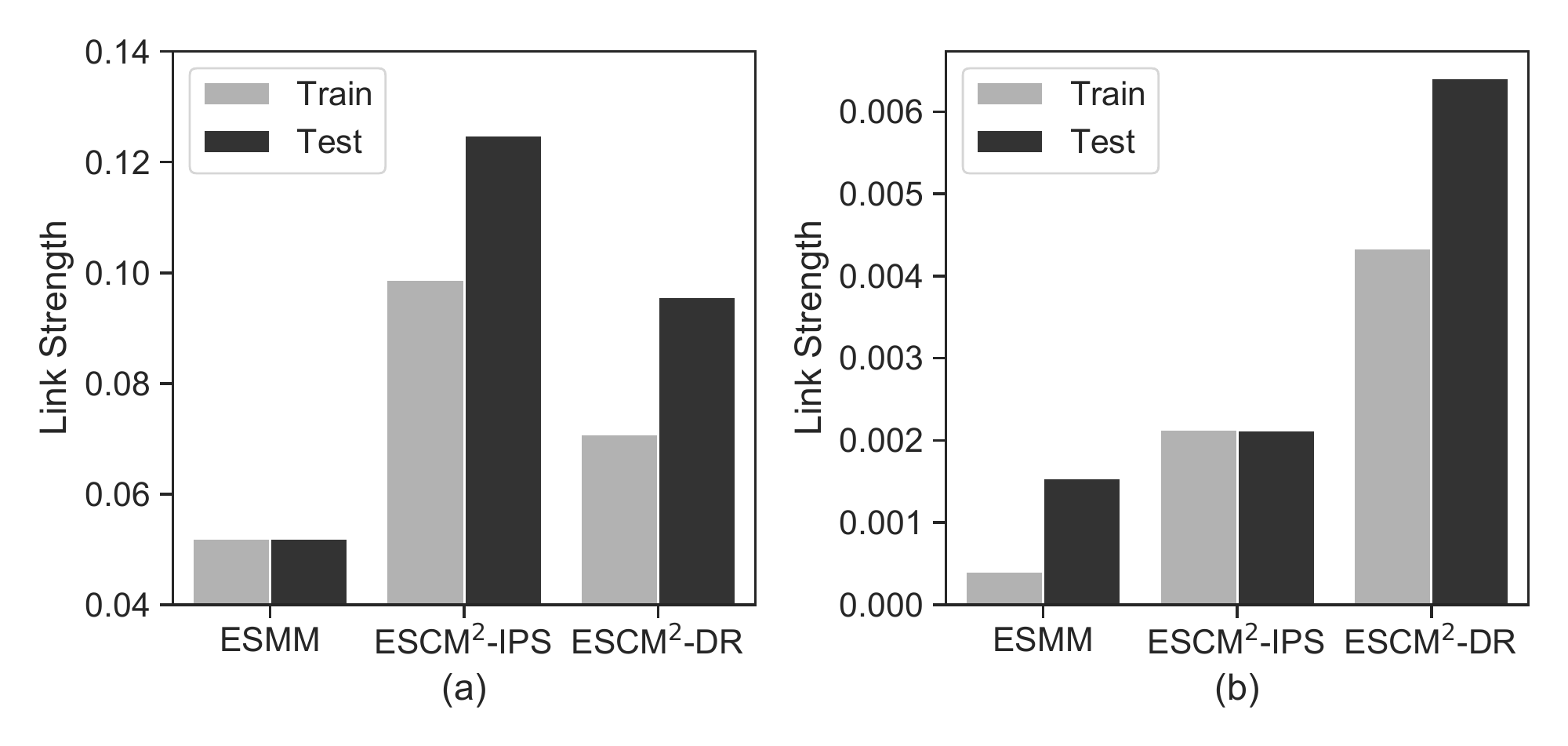}
    \caption{The causality strength of O $\rightarrow$ R, measured by the absolute error of the causal risk ratio to 1 on (a) industrial dataset and (b) Ali-CCP.}
    \label{fig:causality}
\end{figure}

PSM eliminates the confounding effects, which makes it possible to estimate the causal effect through statistical estimators. Causal risk ratio (CRR) is an important metric to estimate the causal effect \cite{whatif}. In short, the closer the CRR is to 1, the weaker the causality is, and vice versa. As such, we model the causality strength as the absolute error between CRR and 1.

\figref{fig:causality} reports the causal strength from the CTR estimates to the CVR estimates. On both datasets, the strengths of ESMM are negligible, which backs up our claim regarding PIP. ESCM$^2$ augments the strength significantly by modeling the causal dependency directly, which supports our arguments in \thmref{thm:anti-pip}.

\subsection{Parameter Sensitivity and Ablation Study}
\begin{figure}
    \centering
\includegraphics[width=\linewidth,trim=20 10 20 10]{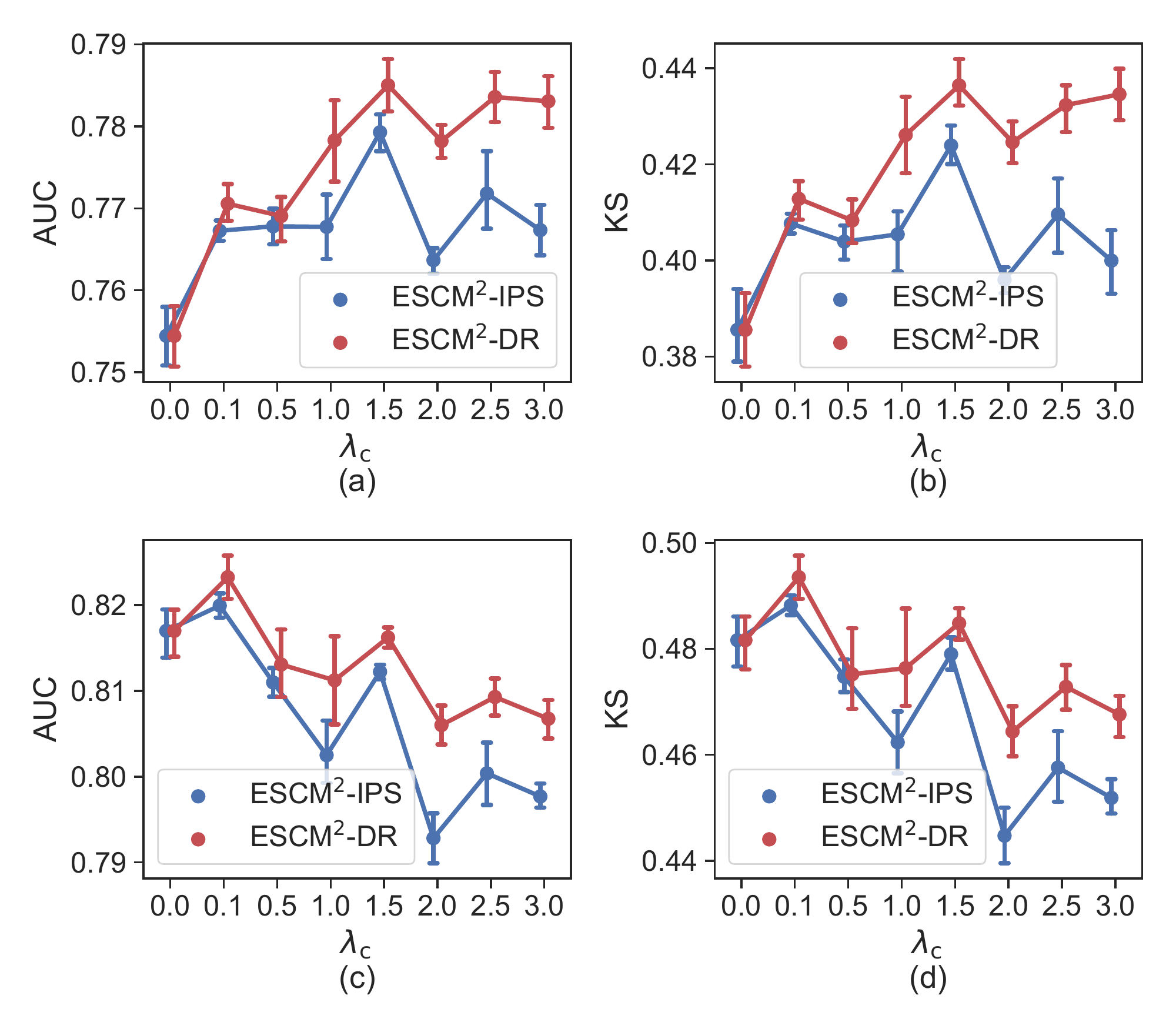}
    \caption{Parameter study of the counterfactual risk weight $\lambda_\mathrm{c}$ on (a-b): CVR estimation; (c-d): CTCVR estimation.}
    \label{fig:lambda_c}
\end{figure}
We discuss two critical hyperparameters in ESCM$^2$, \ie $\lambda_\mathrm{c}$ and $\lambda_\mathrm{g}$ in \eqaref{eq:escm},  which are the weights in the learning objective and influence the final performance significantly. 

In \figref{fig:lambda_c}, we vary $\lambda_\mathrm{c}$ in the range [0,3] to investigate the influence of causal regularization. Specifically, increasing $\lambda_\mathrm{c}$ consistently improves the ranking performance of CVR estimates. For example, the AUC of ESCM$^2$-IPS increases from 0.755 at $\lambda_\mathrm{c}=0$ to approximately 0.78 at $\lambda_\mathrm{c}=1.5$. 
Another observation is that causal regularization also benefits CTCVR estimates, with AUC increasing from 0.817 at $\lambda_\mathrm{c}=0$ to 0.821 at $\lambda_\mathrm{c}=0.1$. However, assigning large weight to CVR risk is detrimental to CTCVR estimates. We speculate that placing a higher focus on CVR risk in a multitask learning framework leads to difficulties in CTR learning, which in turn leads to sub-optimal CTCVR estimation. It might also be due to high variance of IPS and sparsity of clicking data within batches. As such, we generally finetune $\lambda_\mathrm{c}$ within [0,0.1].
\begin{figure}
    \centering
\includegraphics[width=\linewidth,trim=20 10 20 10]{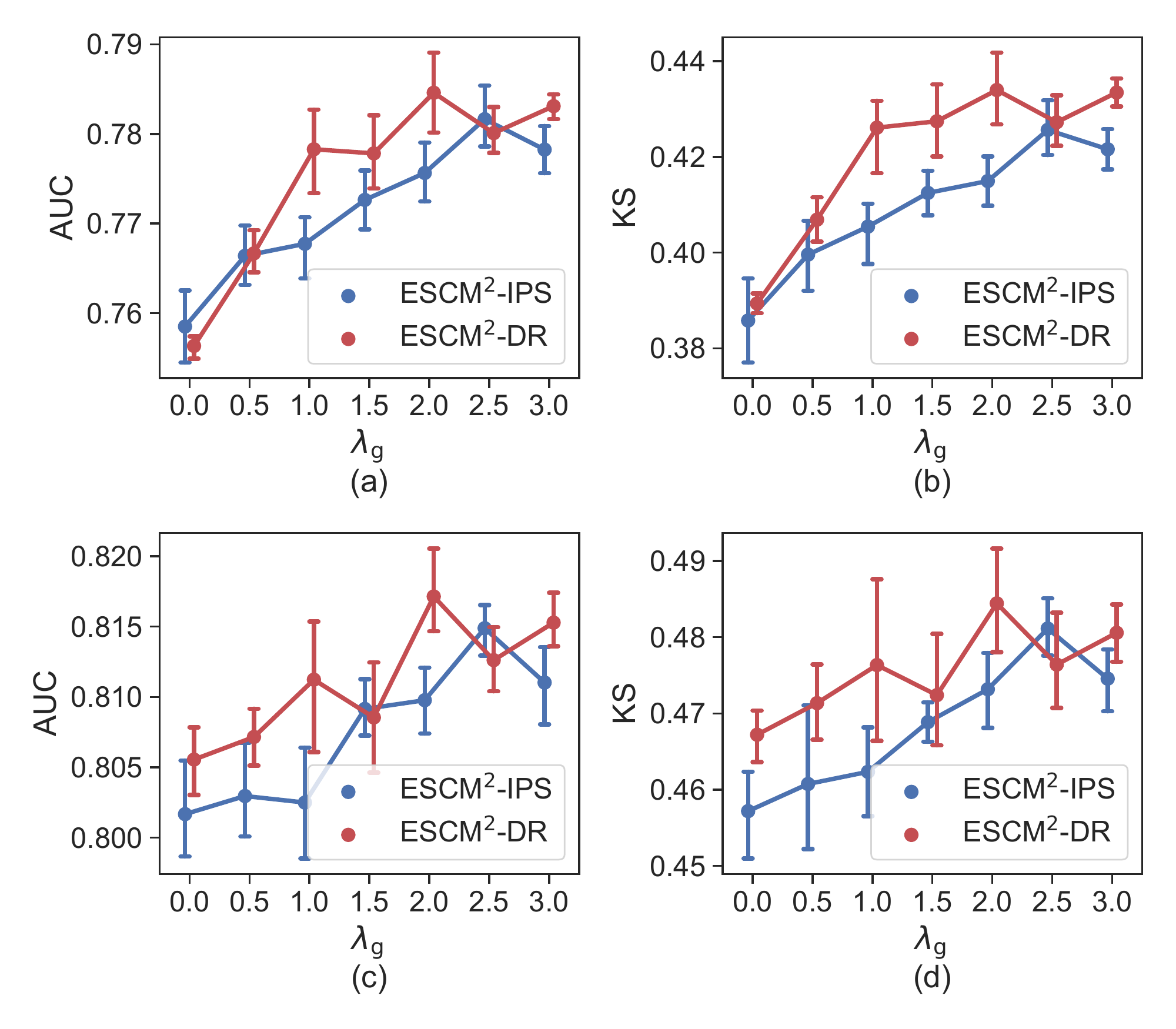}
    % \caption{Effect of global risk weight $\lambda_\mathrm{g}$ on CVR estimation in (a,b) and CTCVR estimation in (c,d).}
    \caption{Parameter study of the global risk weight $\lambda_\mathrm{g}$ on (a-b): CVR estimation; (c-d): CTCVR estimation.}
    \label{fig:lambda_g}
\end{figure}

In \figref{fig:lambda_g}, we vary $\lambda_\mathrm{g}$  to investigate the influence of global risk minimizer. Specifically, within the range [0,3], increasing $\lambda_\mathrm{g}$ consistently improves both CVR and CTCVR estimation. For example, the AUC of ESCM$^2$-IPS increases from 0.758 at $\lambda_\mathrm{g}=0$ to approximately 0.781 at $\lambda_\mathrm{g}=2.5$, and the KS of ESCM$^2$-DR increases from 0.385 at $\lambda_\mathrm{g}=0$ to approximately 0.434 at $\lambda_\mathrm{g}=2.5$. 

%% file: 6_related.tex
\section{Related work}
Multi-task learning with task dependency is a challenging subject in recommender system because dependencies between tasks always make feedback labels MNAR \cite{treatment,mengyue}. Current works in this community can be broadly divided into two groups with respect to the methodology used to address the MNAR problem.

Methods in the first group leverage the sequential pattern of user behaviors to conduct probability decomposition and circumvent the MNAR problem. For example, ESMM \cite{esmm} models CTCVR as the product of CTR and CVR and learns CVR indirectly. ESM$^2$ \cite{esm2} utilizes additional purchase-related post-click behaviors when conducting probability decomposition. GMCM \cite{gmcm} encodes user micro behaviors as graphs and utilizes graph convolutional networks to model interactions. HM$^3$ \cite{hm3} models both micro and macro user behaviors explicitly in a unified deep learning framework.

Methods in the second group aim to build unbiased estimators on biased datasets directly. \citeauthor{treatment} \cite{treatment} proposes to conduct unbiased training and evaluation leveraging causality methodology, which inspires many subsequent research works. \citeauthor{dr} \cite{dr} analyzes the unbiasedness and effectiveness of doubly robust estimator in recommendation system, and \citeauthor{mrdr} \cite{mrdr} extends it to a more robust implementation. \citeauthor{mtlips} \cite{mtlips} proposes to co-learn CTR, CVR and conducted debiasing within a multi-task learning framework. Recently, \citeauthor{dual} \cite{dual} proposes a dual learning framework that simultaneously eliminates the confounding effect in clicked and unclicked data.